\documentclass[10pt,twocolumn,letterpaper]{article}

\usepackage[pagenumbers]{cvpr} 

\usepackage{graphicx}
\usepackage{amsmath}
\usepackage{amssymb}
\usepackage{booktabs}
\usepackage{balance}

\usepackage[pagebackref,breaklinks,colorlinks]{hyperref}

\usepackage[capitalize]{cleveref}

\usepackage[utf8]{inputenc} 
\usepackage{url}            
\usepackage{booktabs}       
\usepackage{amsfonts}       
\usepackage{nicefrac}       
\usepackage{microtype}      
\usepackage{framed,multirow}
\usepackage{amssymb}
\usepackage{latexsym}

\usepackage{microtype}
\usepackage{graphicx}
\usepackage{amsmath}
\usepackage{amssymb}
\usepackage{xcolor}

\usepackage{booktabs}
\usepackage{multirow}
\usepackage{setspace}

\usepackage{pifont}
\usepackage{bm}

\usepackage{wrapfig}
\usepackage{multirow}

\usepackage{arydshln}

\usepackage[accsupp]{axessibility}  

\usepackage{titling}

\usepackage{fancyhdr}
\fancypagestyle{firstpage}{
  \fancyhf{} 
  \fancyhead[C]{To appear in Proceedings of the 2023 IEEE/CVF Conference on Computer Vision and Pattern Recognition (CVPR) \\ The final publication will be available soon.} 
}

\newcommand{\new}[1]{{#1}} 

\newcommand{\red}[1]{{\color{red}{{#1}}}}
\newcommand{\green}[1]{{\color[rgb]{0,0.6,0}{{#1}}}}
\newcommand{\blue}[1]{{\color{blue}{{#1}}}}

\crefname{section}{Sec.}{Secs.}
\Crefname{section}{Section}{Sections}
\Crefname{table}{Table}{Tables}
\crefname{table}{Tab.}{Tabs.}

\def\cvprPaperID{6670} 
\def\confName{CVPR}
\def\confYear{2023}

\begin{document}

\title{Enhancing Deformable Local Features by Jointly Learning \\ to Detect and Describe Keypoints}

\date{}

\author{
Guilherme Potje$^1$ \hspace{0.3in}
Felipe Cadar$^1$    \hspace{0.3in}
André Araujo$^2$    \vspace{0.02in}\\ 
Renato Martins$^{3,4}$   \hspace{0.3in}
Erickson R. Nascimento$^{1,5}$ \vspace{0.1in} \\
$^1$Universidade Federal de Minas Gerais \hspace{0.04in}
$^2$Google Research \\
$^3$Université de Bourgogne \hspace{0.04in} $^4$Université de Lorraine, LORIA, Inria  \hspace{0.04in} $^5$Microsoft \\
{\tt\small \{guipotje,cadar,erickson\}@dcc.ufmg.br, renato.martins@u-bourgogne.fr, andrearaujo@google.com}
}
\maketitle

\begin{abstract}
Local feature extraction is a standard approach in computer vision for tackling important tasks such as image matching and retrieval.
The core assumption of most methods is that images undergo affine transformations, disregarding more complicated effects such as non-rigid deformations. Furthermore, incipient works tailored for non-rigid correspondence still rely on keypoint detectors designed for rigid transformations, hindering performance due to the limitations of the detector. We propose DALF (Deformation-Aware Local Features), a novel deformation-aware network for jointly detecting and describing keypoints, to handle the challenging problem of matching deformable surfaces. All network components work cooperatively through a feature fusion approach that enforces the descriptors' distinctiveness and invariance. 
Experiments using real deforming objects showcase the superiority of our method, where it delivers $8$\% improvement in matching scores compared to the previous best results. Our approach also enhances the performance of two real-world applications: deformable object retrieval and non-rigid 3D surface registration. \new{Code for training, inference, and applications are publicly available at 
\url{verlab.dcc.ufmg.br/descriptors/dalf_cvpr23}.}

\thispagestyle{firstpage} 

\end{abstract}



\section{Introduction}



\begin{figure}[t!]
	\centering
	\includegraphics[width=0.95\columnwidth]{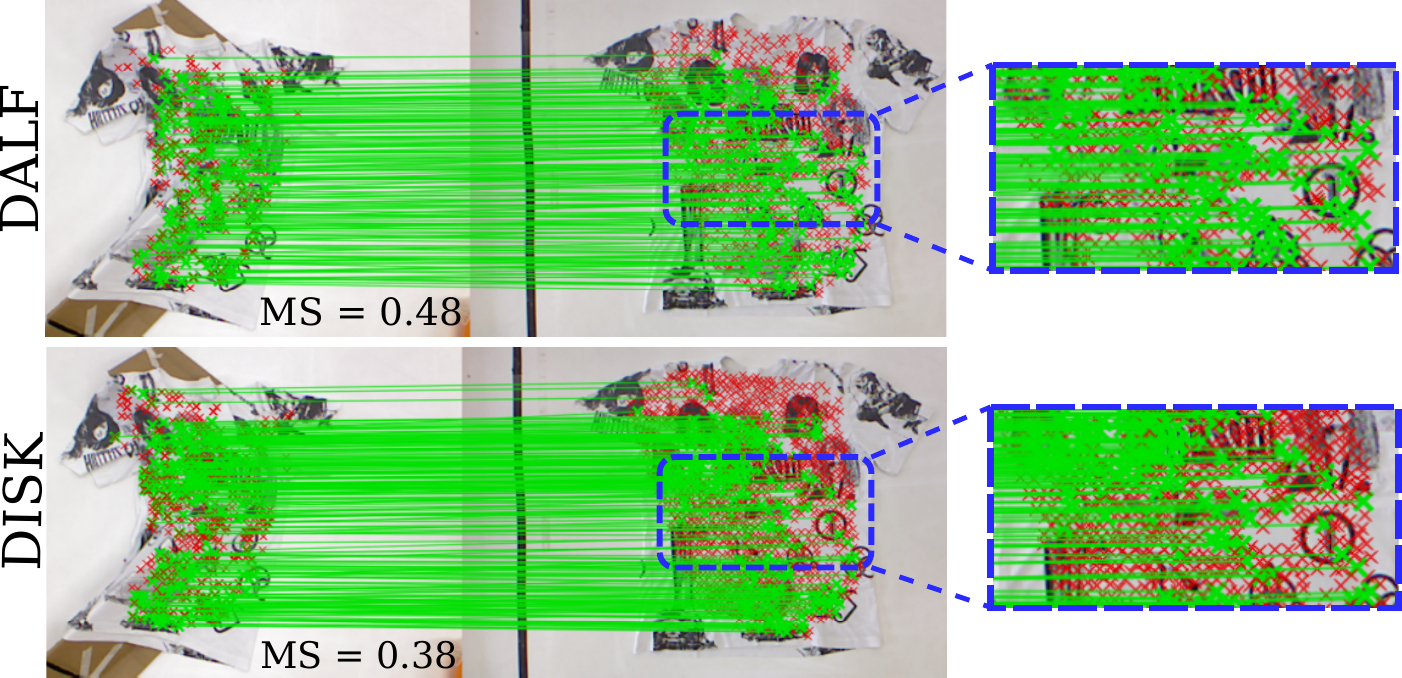}
	\caption{\textbf{Image matching under deformations}. We propose DALF, a deformation-aware keypoint detector and descriptor for matching deformable surfaces. DALF (top) enables local feature matching across deformable scenes with improved matching scores (MS) compared to state-of-the-art, as illustrated with DISK~\cite{cit:DISK}. Green lines show correct matches, and red markers, the mismatches. \vspace{-0.15in}}
 

	\label{fig:teaser}  
\end{figure}

Finding pixel-wise correspondences between images depicting the same surface is a long-standing problem in computer vision. 
Besides varying illumination, viewpoint, and distance to the object of interest, real-world scenes impose additional challenges. The vast majority of the correspondence algorithms in the literature assume that our world is rigid, but this assumption is far from the truth. It is noticeable that the community invests significant efforts into novel architectures and training strategies to improve image matching for rigid scenes~\cite{cit:l2net,cit:hardnet, cit:r2d2, cit:DISK, lift-eccv16, cit:superpoint}, but disregards the fact that many objects in the real world can deform in more complex ways than an affine transformation. 

Many applications in industry, medicine, and agriculture require tracking, retrieval, and monitoring of arbitrary deformable objects and surfaces, where a general-purpose matching algorithm is needed to achieve accurate results. 
Since the performance of standard affine local features significantly decreases for scenarios such as strong illumination changes and deformations, a few works considering a wider class of transformations have been proposed~\cite{simo2015dali, cit:geopatch, cit:deal}.
However, all the deformation-aware methods neglect the keypoint detection phase, limiting their applicability in challenging deformations. Although the problems of keypoint detection and description can be treated separately, recent works that jointly perform detection and description of features~\cite{cit:r2d2, cit:DELG} indicate an entanglement of the two tasks since the keypoint detection can impact the performance of the descriptor. The descriptor for its turn can be used to determine reliable points optimized for specific goals. In this work, we propose a new method for jointly learning keypoints and descriptors robust to deformations, viewpoint, and illumination changes. We show that the detection phase is critical to obtain robust matching under deformations. \cref{fig:teaser} depicts an image pair with challenging deformations, where our method can extract reliable keypoints and match them correctly, significantly increasing matching scores compared to the recent state-of-the-art approach DISK~\cite{cit:DISK}. 

\paragraph{Contributions.}
\textbf{(1)} Our first contribution is a new end-to-end method called DALF (Deformation-Aware Local Features), which jointly learns to detect keypoints and extract descriptors with a mutual assistance strategy to handle significant non-rigid deformations. Our method boosts the state-of-the-art in this type of feature matching by $8$\% using only synthetic warps as supervision, showing strong generalization capabilities. We leverage a reinforcement learning algorithm for unified training, combined with spatial transformers that capture deformations by learning context priors affecting the image;
\textbf{(2)} Second, we introduce a feature fusion approach, \new{a major difference from previous methods} that allows the model to tackle challenging deformations with complementary features (with distinctiveness and invariance properties) obtained from both the backbone and the spatial transformer module. This approach is shown beneficial with substantial performance improvements compared to the non-fused features;
\textbf{(3)} Finally, we demonstrate state-of-the-art results in non-rigid local feature applications for deformable object retrieval and non-rigid 3D surface registration. We also will make the code and both applications publicly available to the community.
\section{Related work}

\noindent\textbf{Keypoint detection.}
Traditional image keypoint detection methods seek to extract repeatable regions in images, \ie, localized points that are stable under different viewing conditions. The classic Harris detector~\cite{harris} employs image derivatives that are used to compute cornerness scores, while one the most used handcrafted detectors SIFT~\cite{lowe2004ijcv}, for instance, detects blobs using the Difference of Gaussians. 
Key.Net~\cite{cit:key.net} showed that it is possible to improve keypoint detection by combining handcrafted filters and learned filters. A recent trend for learning keypoint detection is to couple description and detection in the same pipeline~\cite{lift-eccv16,cit:l2net,cit:r2d2, cit:d2net}, since it is advantageous performance-wise to solve both tasks simultaneously in terms of computation and matching accuracy. In the same direction, our proposed method has a backbone that computes both keypoints and descriptors but at the same time employs a deformation-aware module.

\vspace{0.3cm}
\noindent\textbf{Description of local patches.}
Until recently, detection and description were treated separately. While some works focused on both problems, such as the seminal works of  SIFT~\cite{lowe2004ijcv} and ORB~\cite{rublee2011iccv}, the detection and description were decoupled. SIFT and ORB descriptors employ a handcrafted gradient analysis to extract descriptors with scale and rotation invariance. Recent description approaches based on CNNs~\cite{tfeat-balntas2016, cit:l2net, cit:hardnet, cit:geodesc, cit:beyondcartesian} consume a local patch assuming a pre-defined  keypoint detector. These methods achieved state-of-the-art performance using SIFT keypoints in the image matching benchmark~\cite{jin2020image}. The networks are trained using metric learning~\cite{cit:contrastive, tfeat-balntas2016}. As patch-based methods rely on a pre-defined keypoint detector that may produce keypoints in unreliable or ambiguous regions, noise can be easily introduced because detection and description steps are decoupled. Unlike patch-based methods, our network is trained to produce reliable descriptions and keypoints optimized for non-rigid correspondence. Our description is also enhanced with a fusion strategy that combines complementary features into a single learned feature representation.



\vspace{0.3cm}
\noindent\textbf{Joint detection and description.}
 DELF~\cite{noh2017cvpr} and DELG~\cite{cit:DELG} works demonstrated that coupling the detection and description phases using an attentive mechanism for keypoint selection based on higher-level image semantics can substantially improve retrieval performance. Local feature extraction has been shifting towards learning both detection and description of local features jointly~\cite{lift-eccv16, cit:d2net, cit:r2d2, cit:aslfeat, cit:DISK}. Most methods follow a similar architecture, adopting a fully convolutional network (FCN) layout to produce a dense feature map, where most of the differences between the methods reside in the training scheme and loss design. The most recent describe-and-detect approaches are currently state-of-the-art on standard  benchmarks~\cite{hpatches_2017_cvpr,jin2020image}. 
Differently from existing methods, we design the architecture and training to explicitly handle deformations in a coupled detection and description phase, devising a carefully tailored warper network. 

\vspace{0.3cm}
\noindent\textbf{Deformation-aware methods.}
One of the first proposed image descriptors designed for deformable surfaces is DaLI~\cite{simo2015dali}. DaLI interprets image patches as a local 3D surface, and computes the Scale-Invariant Heat Kernel Signatures~\cite{cit:sihks} of the 3D surface to encode features robust to non-rigid deformations and illumination changes. Despite achieving improved matching performance compared to contemporary works, DaLI suffers from high computational and storage requirements. Similar to DaLI, GeoBit~\cite{nascimento2019iccv} and GeoPatch~\cite{cit:geopatch} descriptors brought ideas from computational geometry to computer vision, leveraging RGB-D images to extract visual features that are geodesic-aware. 
However, these geodesic-aware methods require RGB-D images and are sensitive to noise, significantly restricting their applicability. To remove the need for depth images for estimating geodesic patches, the DEAL descriptor~\cite{cit:deal} implicitly handles deformations from monocular images through a non-rigid warper module. 
The main shortcoming of DEAL is its dependency on existing keypoint detectors, which compromises the descriptor performance due to the lack of equivariance on keypoints locations from most existing detectors in the presence of deformation changes. In contrast, our method learns detection and description in the same framework, achieving significant performance gains.

\section{Methodology}
\begin{figure}[tb!]
	\centering
	\includegraphics[width=0.95\columnwidth]{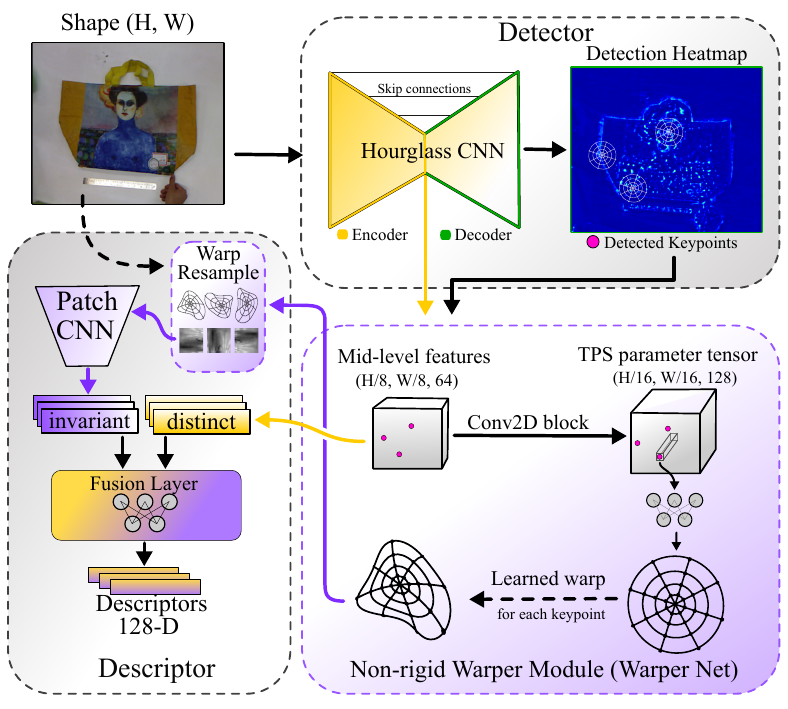}\vspace*{-0.2cm}
	\caption{\textbf{DALF architecture}. Our architecture jointly optimizes non-rigid keypoint detection and description, and explicitly models local deformations for descriptor extraction during training. An hourglass CNN computes a dense heat map providing specialized keypoints that are used by the Warper Net  to extract deformation-aware matches. A feature fusion layer balances the trade-off between invariance and distinctiveness in the final descriptors.
 }
	\label{fig:overview}  \vspace*{-0.5cm}
\end{figure}

DALF jointly learns to detect and describe points robust to non-rigid deformations, in addition to perspective and illumination changes. 
Both detector and descriptor are trained with a cooperative scheme aiming at the invariance of feature representations. Specifically, the keypoint detector is trained using policy gradient, seeking to increase the probability of detections that are both repeatable and reliable; Concurrently, the descriptor extractor learns to undeform and extract discriminative and invariant features from local regions. The model is only trained on synthetic warps, \ie, it does not require expensive human annotation nor pseudo-ground-truth that may contain errors and bias, such as the output of an SfM pipeline that is used in several works~\cite{cit:DISK, cit:superglue, cit:d2net, cit:r2d2, cit:aslfeat}. \cref{fig:overview} outlines the proposed method.


\subsection{Keypoint detector}
The keypoint detection architecture uses a backbone hourglass CNN network $\mathbf{f}(\cdot)$, similar to a U-net~\cite{cit:unet}. This network enables computing a keypoint heat map in the original image resolution efficiently, while also producing mid-level feature representations that are useful to describe the keypoints. We employ three downsampling blocks for the encoder, and three upsampling blocks for the decoder, with skip connections, each having two convolutional layers composed of a 2D convolution followed by ReLU and batch normalization. Let $I \in \mathbb{R}^{h \times w \times c}$ be the input image of size $h \times w$ and $c$ channels, $\mathbf{f}(I)$ outputs two feature maps: mid-level representations $\mathbf{X} \in \mathbb{R}^{h/8 \times w/8 \times d}$ and detection heatmap $\mathbf{H} \in \mathbb{R}^{h \times w}$, where $d$ is the number of features. 

\begin{figure*}[tb!]
	\centering
	\includegraphics[width=0.9\textwidth]{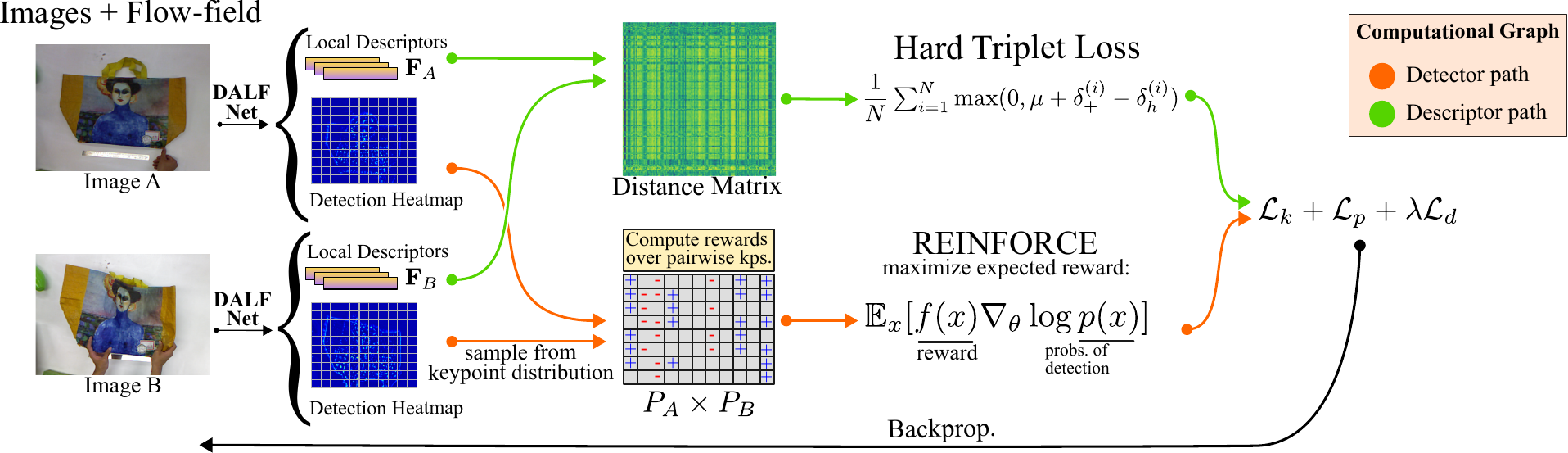}

	\caption{{\bf {Training strategy to learn to detect and describe keypoints aware of deformations.}} DALF network is used to produce a detection heatmap and a set of local features for each image. In the detector path, the heatmaps are optimized via the REINFORCE algorithm considering keypoint repeatability under deformations. In the descriptor path, feature space is learned via the hard triplet loss. A siamese setup using image pairs is employed to optimize the network. 
 Notice that we penalize keypoints that cannot be described accurately during the second training stage; thus, the keypoints and combined descriptors are optimized jointly to be robust to deformations.}\vspace*{-0.4cm}
	\label{fig:method}
\end{figure*}

\paragraph{Keypoint detection in deforming images.} \label{sec:keypoint}

An effective detector must output heatmaps $\mathbf{H} \in \mathbb{R}^{h \times w}$ with high responses in regions that can be matched well in non-rigid scenes containing view and illumination changes. Thus, during the training of the detection branch, we optimize $\mathbf{H}$ using a strategy similar to DISK~\cite{cit:DISK}, but applying only the probabilistic framework to learn the detection heatmap. \new{A key difference compared to DISK is that we enforce the reliability of the detected keypoints by penalizing wrong matches even if the keypoints are repeatable.} A probabilistic approach has several advantages as dealing with the inhering discreteness of sparse keypoint detection, and a simpler and more intuitive loss can be used for better convergence and regularization of the detection heatmap, in contrast to works that require elaborated handcrafted losses~\cite{cit:r2d2, cit:d2net, cit:aslfeat}.

We seek to obtain high responses in confident regions not only for detection but also matching. To solve this problem with policy gradient, we divide the heatmap into a 2D grid of cells (Detection Heatmap in \cref{fig:method}) and consider a set of actions that the network agent can make to select keypoints. Each cell $\mathbf{c}_i$ has $m \times n$ pixels, where the network can learn the probability of detecting a keypoint within each cell. Given an image pair $(A,B)$ of the same scene under different photometric and geometric transformations, and the ground-truth flow-field relating the two images, for each cell $\mathbf{c}_i \in \mathbf{H}$, we consider a probability distribution over the cell $\mathbf{c} \in \mathbb{R}^{m \times n}$. Each logit value within the cell has a probability of being a keypoint. The probability mass function $\mathbf{p}_{\mathbf{c}_i}$ over the cell $\mathbf{c}_i$ is computed by applying the Softmax function.

Therefore, to train the detection branch, we employ the Reinforce algorithm~\cite{cit:policygrad}. During the forward pass of the network, we randomly sample an individual keypoint within each cell $\mathbf{c}_i$ according to the probability mass function $\mathbf{p}_{\mathbf{c}_i}$ alongside the keypoint's spatial coordinates, its probability $p^{i}_{s}$ and its logit $l^{i}_{s}$. Note that each cell can have exactly one keypoint; however, in practice, it is common that low texture and ambiguous regions result in low-quality keypoints that cannot be reliably matched or detected in other images. For that reason, we accept a keypoint proposal from a cell with probability $\sigma(l^{i}_{s})$, where $\sigma$ is the sigmoid activation. This way, the network can learn to filter out unreliable keypoint proposals during training. The final probabilities of detection for the image $I$ is given by the set $P_I = \{ \sigma(l^{i}_{s}) \cdot p^{i}_{s} \}, \forall \mathbf{c}_i \in \mathbf{H}$, such that $\sigma(l^{i}_{s}) > 0.5$ (we only sampled keypoints that has positive values in the heatmap).
Since we want the keypoints to be repeatable, we reward points that can be detected in both images $A$ and $B$. Thus, given the pixel coordinate $\mathbf{p}_A^j \in \mathbb{R}^2$ of detected point $j$ on image $A$, we define the reward function $R(.)$ as follows:
\begin{equation}
 R(\mathbf{p}_A^j) = 
  \begin{cases} 
   1 & \text{if } \exists \mathbf{p}_B^{(.)} \text{ s.t. } \| T(\mathbf{p}_A^j) - \mathbf{p}_B^{(.)} \| < \tau , \\
   0       & \mbox{otherwise},
  \end{cases}
\end{equation}
\noindent where $T(.)$ transforms pixels coordinates of image $A$ to image $B$ according to the ground-truth flow-field, and $\tau$ is a pixel threshold to determine if the detected keypoint in $A$ has a correspondence in image $B$. 

Once we have the set of probabilities $P_A$ and $P_B$, we obtain the gradient of the parameter vector $\theta$ with respect to the expected rewards over all pairwise keypoints ${\mathcal{K} =  P_A \times P_B}$, where $\times$ denotes the cartesian product (see \cref{fig:method}). The gradient ascent is used to maximize the expected sum of rewards:
\begin{equation}
    \nabla_{\theta} \mathbb{E}_{\mathcal{K}}[R(.)] = \sum_{(x,y) \in \mathcal{K}} \nabla_{\theta} (\log p(x;\theta) + \log p(y;\theta)) R(.),
\end{equation}

\noindent where $p(.; \theta)$ denotes the probability of taking that action according to the network parametrized by $\theta$. The variables $x$ and $y$ are the probabilities of detection from a pairwise combination of keypoints. 



During the invariant feature learning stage, after $70\%$ of the training progress, we zero out the reward of the keypoints if their descriptors are unreliable. Details about the penalization term for the keypoints are described in \cref{sec:learning}.

\subsection{Keypoint descriptor} \label{sec:feature_supervision}
We observed that mid-level features extracted from the hourglass encoder do not explicitly model invariance to any kind of deformations, but tend to be highly distinctive on small to moderate photometric and geometric changes, such as varying illumination, and planar warps.
Therefore, it is advantageous to supervise the mid-level features since we obtain informative descriptors with no additional cost at inference. For that, during the first training stage, we bilinearly interpolate the feature maps $\mathbf{X}$ at the detected keypoint positions to obtain a feature vector $\mathbf{f}_d \in \mathbb{R}^{D}$ for each keypoint coordinate. 
Let $\mathbf{F}_A \in \mathbb{R}^{N\times D}$ and $\mathbf{F}_B \in \mathbb{R}^{N\times D}$ be matrices of $N$ L2-normalized feature vectors of corresponding descriptors $\mathbf{f}_d$ extracted by the hourglass decoder at keypoint positions, from images $A$ and $B$, and ${\mathbf{D}_{N\times N} = \sqrt{2 - 2\mathbf{F}_A\mathbf{F}_B^T}}$ the distance matrix. To optimize the descriptors' embedding space, we employ the hard mining strategy~\cite{cit:hardnet} in the matrix $\mathbf{D}$ and minimize the margin ranking loss:
\begin{equation} \label{eqn:hardloss}
\mathcal{L}_d\left(\delta_{+}^{(.)}, \delta_{h}^{(.)} \right) = \dfrac{1}{N} \sum_{i=1}^{N} \max(0, \mu + \delta_{+}^{(i)} - \delta_{h}^{(i)}),
\end{equation}
\noindent where $\mu$ is the margin, $\delta_{+} =  \left\Vert \mathbf{F}(p) - \mathbf{F}(p') \right\Vert_2$ is the distance between the corresponding features, and $\delta_{h} =  \left\Vert \mathbf{F}(p) - \mathbf{F}(h) \right\Vert_2$ is the distance to the hardest negative\footnote{The hardest negative example $h_{i}^{j}$ for each row $j$ of $\mathbf{D}_{N\times N}$ is computed as $\min{D_j}, i = j = \{1, ..., N\} \text{ s.t. } i \neq j$.} sample in the batch.


\subsection{Non-rigid warper module}
CNNs' translation equivariance property makes local descriptors invariant to image translation, and multi-scale strategies increase the robustness of description extraction to in-plane scale changes~\cite{cit:r2d2, cit:DISK, cit:d2net}. However, when non-rigid deformations arise, the local texture can significantly change in appearance, introducing matching ambiguities. DEAL~\cite{cit:deal} demonstrates that thin-plate-splines (TPS), coupled with spatial transformers, can be used to model deformations for the task of local feature description. Inspired by DEAL, we adopt a TPS deformation to learn local invariance to non-rigid transformations affecting the patches.

\vspace{-0.1in}
\paragraph{Spatial transformer network.}
We use the mid-level features from the backbone network to learn the parameters of the TPS with little additional overhead. The TPS parameter tensor $\mathbf{M}_{\theta} \in \mathbb{R}^{h/16 \times w/16 \times 2d}$ contains an intermediate representation useful to estimate a local non-rigid transformation for a keypoint. To obtain the parameter vector used in the TPS equation, first, we bilinearly interpolate a feature vector from $\mathbf{M}_{\theta}$ at the spatial position of the keypoint, obtaining an intermediate parameter vector $\in \mathbb{R}^{2d}$. Then, an MLP is used to estimate the parameter vector $\mathbf{\mu}_{\theta}$ that is used in the TPS transformation. The parameter vector $\mathbf{\mu}_{\theta}$ encodes the affine matrix $\mathbf{A} \in \mathbb{R}^{2 \times 3}$, and the non-rigid components $\mathbf{w}_k \in \mathbb{R}^2$ separately representing offsets from the affine component. Given a homogeneous 2D point $\mathbf{q} \in \mathbb{R}^3$, weight coefficients and control points $\mathbf{c}_k$, $\in \mathbb{R}^2$, we use the parameters contained in $\mathbf{\mu}_{\theta}$ to apply the TPS transformation to a fixed polar grid centered at the keypoint:
\begin{equation} \label{eqn:tps}
    \mathbf{p} = \mathbf{A} \mathbf{q} + \sum_{k=1}^{n_c}{ \rho(\| \mathbf{q} - \mathbf{c}_k\|^2) \mathbf{w}_k },
\end{equation}
\noindent where $n_c$ is the number of control points, $\mathbf{q}$ is a normalized spatial image coordinate from the fixed polar grid around the keypoint, and $\mathbf{p}$ is its transformed coordinate. \cref{fig:overview} (Warper Net) shows the patch warping and sampling step. Since we are using the TPS Radial Basis Function, $\rho=r^2 \log{r}$ is used. After the polar grid is transformed, a differentiable bilinear sampler~\cite{cit:stn} is used to obtain the transformed image patch that is used by a CNN similar to L2-Net architecture~\cite{cit:l2net} to compute the invariant feature vector, supervised by the margin ranking loss (\cref{eqn:hardloss}). In our implementation, a major difference from the original L2-Net is that in the last convolutional block, we add an average pooling in the axis respective to the angular axis in the polar patch, attaining full rotation invariance. 

\subsection{Feature fusion layer}
Distinctiveness and invariance are two desired attributes of a local feature descriptor. While invariance is vital for tasks that handle large appearance changes, such as rotation and scale, it usually implies distinctiveness loss~\cite{cit:varma2007}.
By considering two complementary features, the distinctive ones coming from the backbone network with a larger receptive field but more sensitive to strong geometric transformations, and invariant features coming from the warper module that are robust to deformation and rotation by design, we propose to incorporate both information by a feature fusion step. 

The fusion is performed by an attention-based MLP that predicts weight coefficients. The two descriptor vectors are first concatenated and forwarded to the Fusion Layer as depicted in \cref{fig:overview}. Then, the concatenated descriptors are weighted by the predicted attention weights and L2-normalized to produce the final descriptor. During training, we optimize each descriptor loss individually and the loss of the fused descriptors simultaneously to enforce the network to learn how to fuse the feature vectors to achieve a better feature representation. In the experiments, we demonstrate that combining the features allows the final descriptor to handle strong image transformations while maintaining its distinctiveness.

\subsection{Training strategy and model optimization}\label{sec:learning}

\paragraph{Stage-wise training.}
 During experiments, we observed that training the network end-to-end in a single phase causes the model to focus on the invariant features and ignores the distinctive features coming from the backbone, even when re-weighting the loss terms. 
 To solve the issue, we perform a two-stage training. During the first training stage, we only train the backbone network. 
 The backbone features have a larger receptive field and higher-level semantics compared to the Warper Net features but with less invariance to rotation and low-level deformations. In the second training phase, the Decoder, Warper Net \new{and Fusion Layer} are optimized, where the final feature representation is optimized considering both representations through the fusion step.
 Moreover, the decoder of the network is further refined and encouraged to detect keypoints that are optimal for the fused descriptors. 

\vspace*{0.2cm}\noindent\textbf{Final loss.}
For the detection branch, we define the keypoint loss as $\mathcal{L}_{k} = - \mathbb{E}_{\mathcal{K}}[R(.)]$ and add a regularization term for all the detected keypoints during training $\mathcal{L}_{p} = - \sum_{x} \log p(x) \cdot c$, where $c$ is a small negative constant to discourage the network from detecting low-quality points. We employ the margin ranking loss described in \cref{sec:feature_supervision} for all the descriptor vectors computed by our network. 
The final loss is then computed as $\mathcal{L} = \mathcal{L}_{k} + \mathcal{L}_{p} + \lambda \mathcal{L}_{d}$, where $\lambda$ is a weight term to balance the magnitude between the triplet loss and the policy-gradient losses.

\section{Experiments}

\paragraph{Training and implementation details.} We developed a carefully designed synthetic data generation pipeline to create plausible non-rigid deformations of surfaces to supervise the training. We perform photometric and geometric changes to real images obtained from large-scale Structure-from-Motion datasets~\cite{cit:1dsfm}. We use only the raw images and do not use any information about correspondences or annotated labels. During the training, we add random photometric changes, random homographic projection, and random TPS warps to obtain ground-truth dense flows between image pairs depicting the same surface. The training starts with easier samples and progressively becomes more difficult, achieving the hardest difficulty at $60\%$ of training iterations. 
In the experiments, we use the following hyperparameter values: $\mu=0.5$ in the triplet loss; pixel threshold $\tau=1.5$; $\lambda=0.005$ to balance the loss terms; keypoint penalization $c=-7e^{-5}$; cell size $m = n = 8$ pixels; and the number of control points  $n_c = 64$. As detailed in \cref{sec:learning}, we perform a two-stage training. Gradient accumulation was used for four forward passes before updating the weights. We trained the network for $80{,}000$ iterations in the first stage and $100{,}000$ iterations in the second stage. 
During inference, we use a non-maximal suppression of size $3 \times 3$ pixels in order to extract the keypoint coordinates from $\mathbf{H}$. 
Our network is implemented on PyTorch, has about $1M$ trainable parameters, and takes $48$ hours to train on a GeForce GTX Titan X GPU.

\vspace{0.2cm}\noindent\textbf{Baselines and evaluation metrics.} 
We compare our method with several patch-based descriptors~\cite{rublee2011iccv, cit:alahi2012freak, daisy, nascimento2012iros, tfeat-balntas2016, cit:beyondcartesian, cit:sosnet}, using the same set of SIFT~\cite{lowe2004ijcv} keypoints following the protocol of the image matching benchmark~\cite{jin2020image}. \new{We also perform tests with a detector suitable for non-rigid correspondence~\cite{cit:welerson2022} coupled with the deformation-aware descriptor DEAL~\cite{cit:deal}.} Finally, we also include in the comparison the state-of-the-art detect-and-describe methods~\cite{cit:d2net, cit:lfnet, lift-eccv16, cit:superpoint, cit:r2d2, cit:aslfeat, cit:DISK}. For each evaluated method, we detect the top $2{,}048$ keypoints and match the descriptors using nearest neighbor search. In addition to the standard comparison, we include as the gold standard for image matching SuperPoint~\cite{cit:superpoint} with the SuperGlue~\cite{cit:superglue} matcher, which holds the state-of-the-art for stereo and multi-view camera registration~\cite{jin2020image} assuming rigid scenes. As shown in \cref{table:matching_score}, the methods are divided into three categories: (i) methods that only require RGB input (\textit{RGB}) in contrast to methods that require additional information such as depth; (ii) Detect \& Describe (\textit{D\&D}) methods that provide both detection and description jointly within a single pipeline; and (iii) Deformation-Aware (\textit{D-A}) methods, which take into account deformation when computing the descriptors. Notice that a method may fulfill multiple categories simultaneously.

We used the Matching Scores (MS)~\cite{cit:matchscore} to evaluate the matching performance of both the detected keypoints and descriptors. Given a ground-truth transformation and a threshold in pixels, we compute the set of correct correspondences $\mathbf{S}_{gt}$ and obtain the score for an image pair $(i,j)$ as $MS = |\mathbf{S}_{gt}|/\min(|keypoints_i|, |keypoints_j|)$. In addition, the mean matching accuracy (MMA) is also reported, which focuses on the accuracy of the descriptors to match the keypoints that were successfully detected on both images under the threshold denoted as the set $\mathbf{K}_{gt}$, and is computed as $MMA =|\mathbf{S}_{gt}|/|\mathbf{K}_{gt}|$. \new{Additional results regarding repeatability of keypoints can be found in the supplementary material.} To conduct the evaluation, we adopt two existing datasets of deformable objects~\cite{cit:geopatch,wang2019deformable}.

\begin{table*}[tb!]
	
	\centering
	\caption{{\bf Performances using top $2{,}048$ keypoints}. The \textit{RGB} methods only require color images. \textit{D\&D} methods perform joint detection and description. The deformation-aware methods are shown as \textit{D-A}. The mark $^{*}$ indicates the use of a learned matcher. Best in bold and second-best underlined. The mean was calculated with full-precision values \new{by averaging the scores of all 833 image pairs} before rounding. }
	\label{table:matching_score}
	\resizebox{0.92\linewidth}{!}{%
		
		\begin{tabular}{@{}ccclccccc@{}}
			\toprule 
			 \multirow{3}{*}{\textit{RGB}} & \multirow{3}{*}{\textit{D\&D}} & \multirow{3}{*}{\textit{D-A}} & \multirow{3}{*}{{\bf Method}} & \multicolumn{4}{c}{{\bf Datasets: $833$ pairs total}  -- MS / MMA @ 3 pixels $\uparrow$} &  \multirow{3}{*}{\bf Mean} \\
			\cmidrule(lr){5-8}
			 & & & & \textit{Kinect 1}~\cite{cit:geopatch} & \textit{Kinect 2}~\cite{cit:geopatch} & \textit{DeSurT}~\cite{wang2019deformable}  & \textit{Simulation}~\cite{cit:geopatch} & \\
			\cmidrule(r){1-3}\cmidrule(lr){4-4}\cmidrule(l){5-8} \cmidrule(l){9-9}
			\phantom & \phantom & \phantom & BRAND~\cite{nascimento2012iros} 		& $0.17$ / $0.34$ & $0.22$ / $0.49$ & $0.14$ / $0.33$ & $0.04$ / $0.09$ & $0.16$ / $0.34$ \\
			\checkmark & \phantom & \phantom & ORB~\cite{rublee2011iccv} 		& $0.19$ / $0.38$ & $0.25$ / $0.55$ & $0.18$ / $0.40$ & $0.14$ / $0.30$ & $0.20$ / $0.43$  \\	
			\checkmark & \phantom & \phantom & DAISY~\cite{daisy} 		& $0.23$ / $0.47$ & $0.29$ / $0.62$ & $0.16$ / $0.37$ & $0.19$ / $0.39$ & $0.22$ / $0.48$  \\			
			\checkmark & \phantom & \phantom & FREAK~\cite{cit:alahi2012freak} 		& $0.24$ / $0.49$ & $0.33$ / $0.72$ & $0.16$ / $0.38$ & $0.15$ / $0.31$ & $0.23$ / $0.51$  \\
			\checkmark & \phantom & \phantom & TFeat~\cite{tfeat-balntas2016} 		& $0.25$ / $0.50$ & $0.28$ / $0.61$ & $0.21$ / $0.48$ & $0.29$ / $0.63$ & $0.26$ / $0.56$  \\	
			\checkmark & \phantom & \phantom & Log-Polar~\cite{cit:beyondcartesian}	& $0.28$ / $0.58$ & $0.30$ / $0.65$ & ${0.23}$ / ${0.54}$ & $0.22$ / $0.49$ & $0.26$ / $0.57$  \\
			\checkmark & \phantom & \phantom & SOSNet~\cite{cit:sosnet} &  $0.17$ / $0.34$ & $0.25$ / $0.55$ & $0.17$ / $0.38$ & $0.26$ / $0.57$ & $0.22$ / $0.47$ \\
			\hdashline \noalign{\vskip 0.5ex}
			\checkmark & \checkmark & \phantom & LF-Net~\cite{cit:lfnet} & $0.44$ / $0.40$ & $0.51$ / $0.43$ & $0.28$ / $0.77$ & $0.21$ / $0.74$ & $0.36$ / $0.59$ \\
			\checkmark & \checkmark & \phantom & LIFT~\cite{lift-eccv16} & $0.09$ / $0.57$ & $0.16$ / $0.65$ & $0.08$ / $0.52$ & $0.13$ / $0.73$ & $0.12$ / $0.62$ \\	
			\checkmark & \checkmark & \phantom & D2-Net~\cite{cit:d2net} & $0.20$ / $0.50$ & $0.23$ / $0.82$ & $0.14$ / $0.47$ & $0.11$ / $0.30$ & $0.17$ / $0.57$ \\
			\checkmark & \checkmark & \phantom & SuperPoint~\cite{cit:superpoint} & $0.45$ / $0.74$ & $\underline{0.54}$ / $0.85$ & $0.39$ / $0.68$ & $0.18$ / $0.34$ & $0.41$ / $0.69$ \\
			\checkmark & \checkmark & \phantom & R2D2~\cite{cit:r2d2} & $0.17$ / $0.36$ & $0.25$ / $0.59$ & $0.14$ / $0.32$ & $0.06$ / $0.16$ & $0.17$ / $0.39$ \\
			\checkmark & \checkmark & \phantom & ASLFeat~\cite{cit:aslfeat} & $0.31$ / $0.58$ & $0.39$ / $0.69$ & $0.28$ / $0.53$ & $0.19$ / $0.35$ & $0.31$ / $0.56$ \\
			\checkmark & \checkmark & \phantom & DISK~\cite{cit:DISK} & $\underline{0.53}$ / $\underline{0.76}$ & $0.52$ / $0.81$ & $\underline{0.44}$ / $0.61$ & $0.26$ / $0.34$ & $\underline{0.45}$ / $0.66$ \\
			\checkmark & \checkmark & \phantom & SuperGlue$^{*}$~\cite{cit:superglue} & $0.40$ / $0.66$ & $\textbf{0.62}$ / $\textbf{0.99}$ & $0.39$ / $\underline{0.68}$ & $0.23$ / $0.43$ & $0.44$ / $0.74$ \\
			\hdashline \noalign{\vskip 0.5ex}
			\phantom & \phantom & \checkmark & GeoBit~\cite{nascimento2019iccv} 		& ${0.31}$ / $0.65$ & ${0.35}$ / $0.77$ & $0.20$ / $0.47$ & ${0.32}$ / ${0.71}$ & ${0.30}$ / $0.66$  \\
			\phantom & \phantom & \checkmark & GeoPatch~\cite{cit:geopatch} 		& ${0.32}$ / $0.66$ & ${0.35}$ / $0.80$ & $0.26$ / $0.60$ & $\underline{0.39}$ / $\textbf{0.86}$ & ${0.33}$ / $0.73$  \\
			\checkmark & \phantom & \checkmark & DaLI~\cite{simo2015dali} 		& $0.25$ / $0.51$ & ${0.35}$ / $0.76$ & $0.21$ / $0.48$ & $0.10$ / $0.22$ & $0.25$ / $0.54$  \\
			\checkmark & \phantom & \checkmark & SIFT + DEAL~\cite{cit:deal}& $0.33$ / $0.68$ & $0.38$ / $0.85$ & $0.27$ / $0.63$ & $0.36$ / $\underline{0.80}$ & $0.34$ / $\underline{0.75}$  \\
			\checkmark & \phantom & \checkmark & \new{Det.~\cite{cit:welerson2022} + DEAL} & $0.44$ / $0.74$ & $0.49$ / $0.82$ & $0.33$ / $0.64$ & $0.31$ / $0.74$ & $0.40$ / $0.74$  \\
			\midrule	
			\checkmark & \checkmark & \checkmark & DALF (ours) 	& $\textbf{0.54}$ / $\textbf{0.82}$ & $\textbf{0.62}$ / $\underline{0.90}$ & $\textbf{0.49}$ / $\textbf{0.73}$ & $\textbf{0.42}$ / $0.69$ & $\textbf{0.53}$ / $\textbf{0.80}$  \\
			\bottomrule 
		\end{tabular}
		
	}
	
\end{table*}


\subsection{Real-world benchmarking}

\paragraph{Comparison with the state-of-the-art.} \cref{table:matching_score} shows the MS and MMA scores achieved by all compared methods. DALF outperformed all descriptors on average in both MS and MMA metrics, including the methods that use additional depth information to extract deformation-invariant features, improving the state-of-the-art in $8\%$ p.p. in matching scores. Moreover, our method shows promising generalization properties to real deformations. DISK achieved the second-best results in MS, but the MMA indicates that its descriptors are more sensitive to non-rigid deformations. DEAL displays the second-best results in MMA thanks to its deformation-aware module but has poor performance on the MS score. It is noteworthy that DEAL relies on SIFT keypoints, which are not designed for non-rigid transformations.

SuperGlue, which is comprised of SuperPoint descriptors matched with a graph neural network, showed good performance, but the drop in the scores compared to the top methods is noticeable as SuperPoint and SuperGlue do not explicitly model scene deformations. We emphasize that our method can be easily coupled and trained with a learned matcher such as SuperGlue. 
We report the SuperGlue results using the outdoor pretrained weights since we observed that the outdoor weights peformed better in all datasets.
All the other methods achieve significantly worse scores due to their inability to cope with stronger deformations alongside illumination and affine transformations.




\begin{figure}[b!]
	\centering
    \begin{tabular}{cc}
         		\includegraphics[width=0.46\columnwidth]{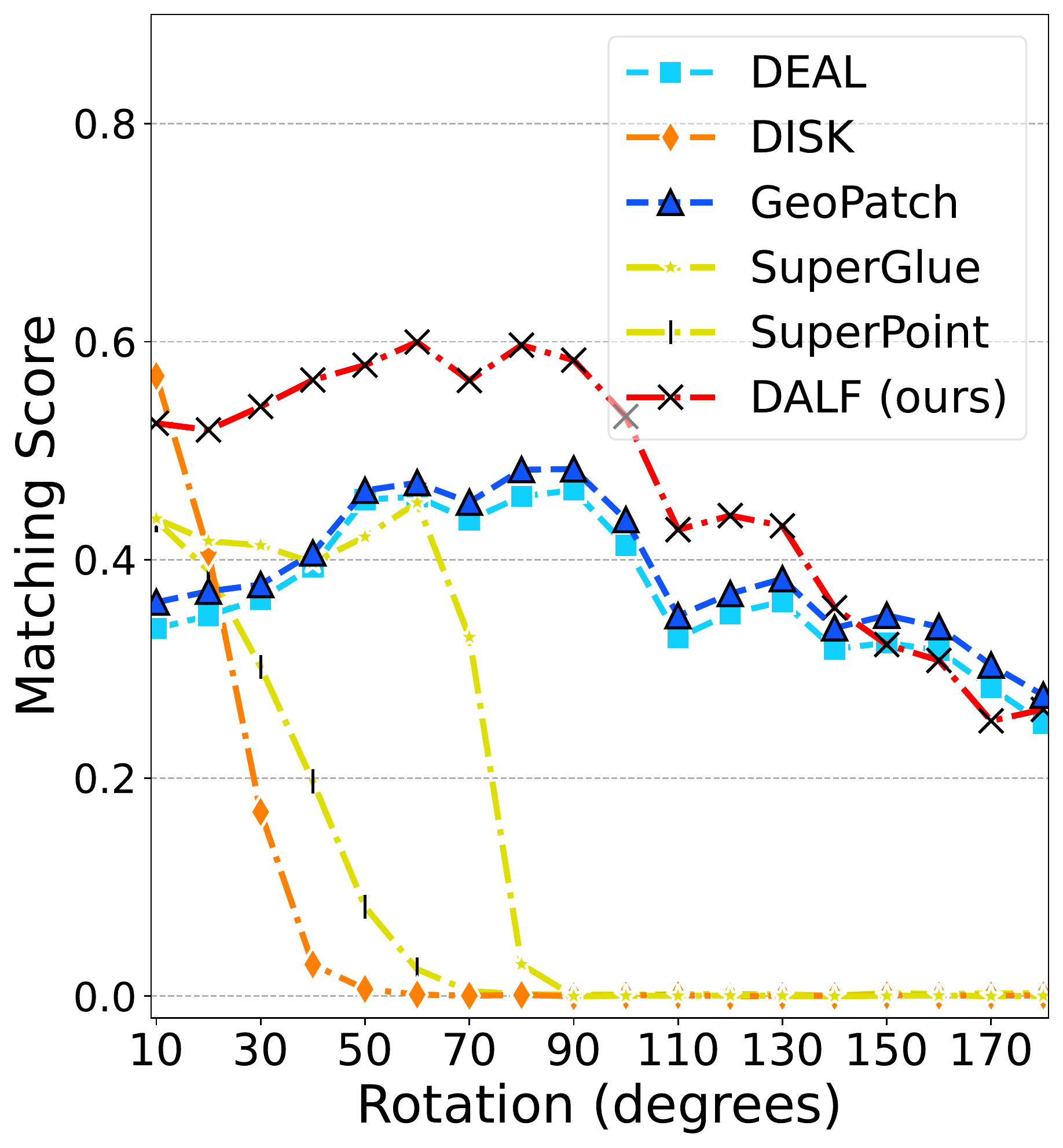} &
              \includegraphics[width=0.46\columnwidth]{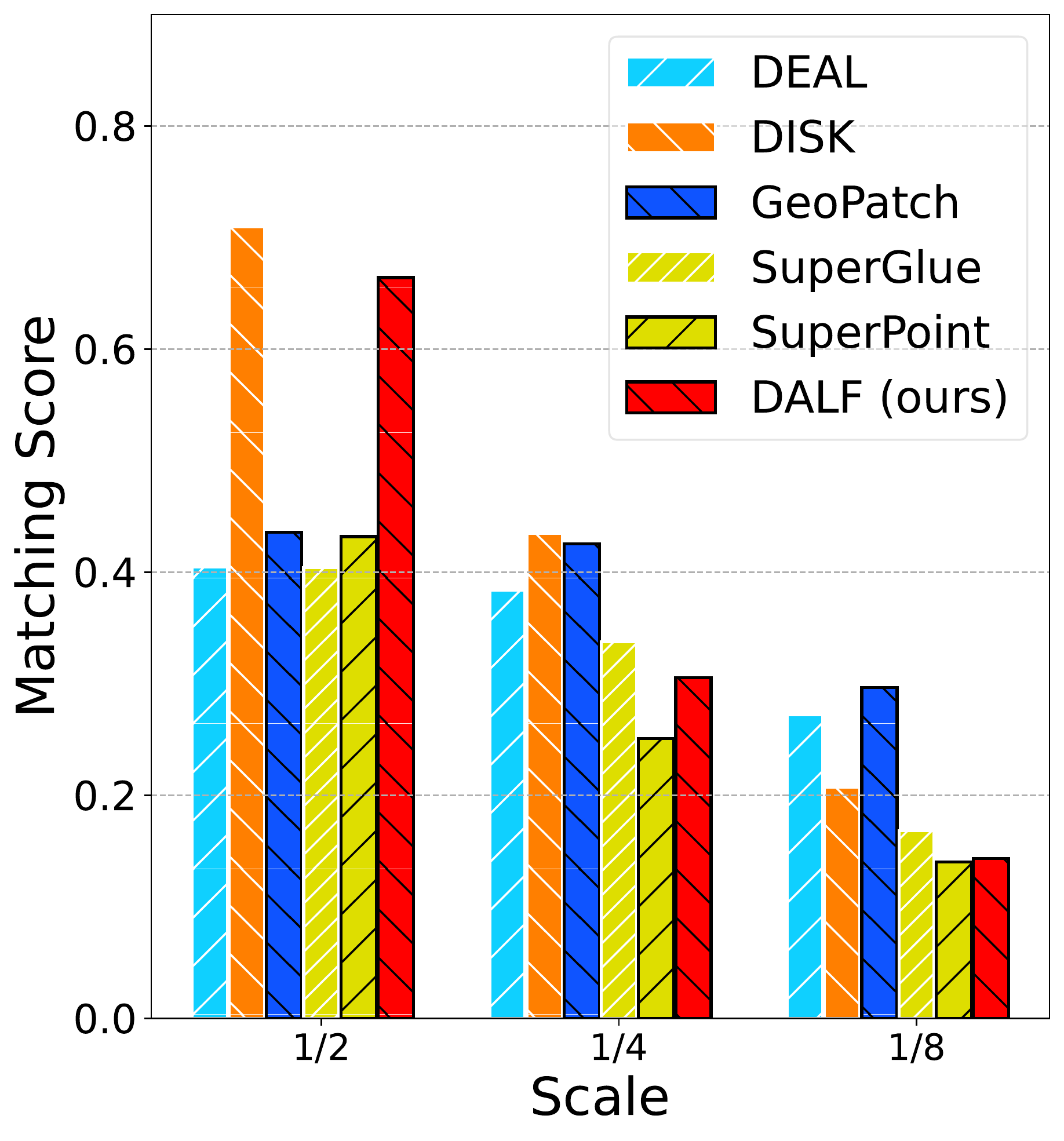}
    \end{tabular}
\vspace{-0.4cm}
	\caption{{\textbf{Invariance to rotation \& scale.} We evaluate the matching performance of the methods under rotation and scale changes between image pairs from the \textit{Simulation} dataset.} The objects are simultaneously deforming, rotating and scaling in image space.}
	\label{fig:invariance-exp}

\end{figure}



\vspace*{0.1cm}\noindent\textbf{Rotation and scale robustness.} 
Aside from deformations, in-plane rotations and scale changes are two important geometric transformations. Thus, we conduct a study using the Simulation sequences from~\cite{cit:geopatch} containing challenging rotation and scale changes.
\cref{fig:invariance-exp} clearly indicates that our method holds the best invariance to image in-plane rotations in addition to deformation changes compared to the five stronger competitors. Our technique also displays considerable robustness to scale changes, outperforming SuperPoint and providing a similar level of robustness of SuperGlue.

\begin{table}[t!]
	\centering
	\caption{{\bf Ablation.} Performance of our method when considering different network components and training strategies.}
     \resizebox{0.93\columnwidth}{!}{%
	\begin{tabular}{cccccr}
		\toprule      
		\textbf{Distinct} & \textbf{Invariant} & \textbf{2-Stage} & \textbf{Attn.} & \textbf{$\uparrow$} MS / MMA  \\ 
		\toprule
		\checkmark & \phantom{n} & - & - & $0.48$ / $0.72$ \\
		\phantom{n} & \checkmark & - & - & $0.51$ / $0.78$\\
		\checkmark & \checkmark & \phantom{n} & \phantom{n} &  $0.53$ / $0.79$\\
		\checkmark & \checkmark & \checkmark & \phantom{n} &  $0.53$ / $0.78$\\
		\checkmark & \checkmark & \checkmark & \checkmark &  $0.53$ / $0.80$\\
		\bottomrule		
	\end{tabular}    
    }
	\label{table:ablation}
\end{table}


		


\vspace*{0.2cm}\noindent\textbf{Time efficiency.} DALF is one of the most time efficient methods among the joint detection and description architectures. While our method runs at $9$ FPS, DISK runs at $5$ FPS and R2D2 at $2$ FPS in an NVIDIA GeForce RTX 3060 GPU to extract $2{,}048$ keypoints from $1024 \times 768$ images. 

\subsection{Ablation study} 

Our ablation study comprises five different configurations of our method: (i) using the U-net backbone only without the non-rigid warper module, which is similar to DISK excepting the descriptor loss term; (ii) computing the descriptors using the non-rigid warper module only; (iii) fusing the invariant and distinct features from the non-rigid warper and backbone respectively; (iv) perform a stage-wise training where the backbone is optimized first and the non-rigid warper second, and finally (v) we perform stage-wise training with an additional attention layer to fuse the invariant and distinctive descriptors instead of simple concatenation.

From \cref{table:ablation}, we can observe that the non-rigid warper contributes significantly to achieving more accurate matches when compared to using a convolutional backbone alone. Furthermore, by fusing the features, it is possible to obtain an improved descriptor that is both invariant and distinct with complementary properties. The two-stage training provides similar matching scores and slightly reduced mean accuracy compared to end-to-end training. Still, \new{according to a more detailed analysis available in the supplementary material}, we observed that it is beneficial to perform stage-wise training. The invariant part tends to dominate the distinctive part during training, rendering the distinct part less useful in practice, which is not desired for applications needing more distinct features, such as image retrieval, and datasets without significant deformations. Finally, we test if an attention-based fusion layer can deliver better results than concatenating the descriptors in the fusion step. According to the results, it is possible to slightly increase the accuracy even further with a negligible cost in computation. Thus, we choose the model with the stage-wise training as the final architecture.

\vspace*{0.1cm}\noindent\textbf{Limitations.} Although our network can improve overall scores by learning keypoints and deformation-aware features, estimating the deformation parameters from a single image is an  ambiguous problem. Therefore, physical deformations may make textures similar for different objects, harming our method's performance. Nevertheless, the learned deformations demonstrate good generalization properties to real deformations due to the powerful combination of the keypoint extractor, warper module, and feature fusion steps. 

\begin{figure}[tb!]
	\centering
	\includegraphics[width=0.99\columnwidth]{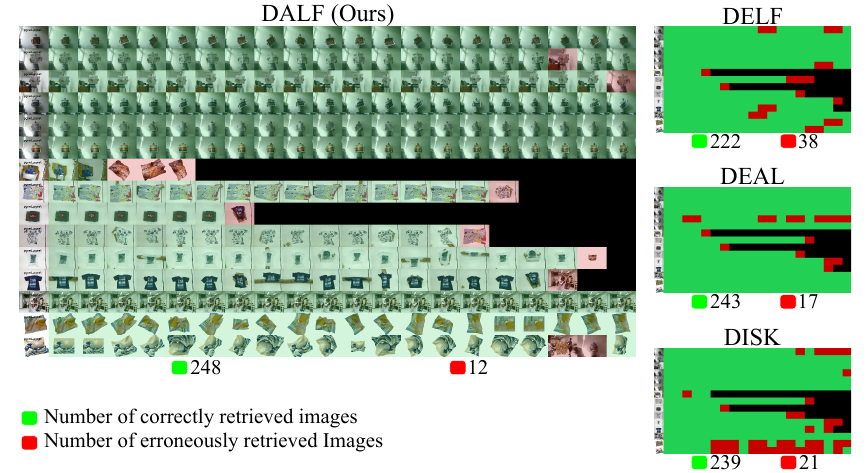}
	
    \caption{{\bf Deformable object retrieval.} Our method has the best result in retrieving images of deformed objects. The first column shows the image queries, and the rows show the results from different queries. Green images correspond to the same object as the query, and red images do not correspond. Black squares imply no more objects are available for that query.} \vspace{-0.12in}
	\label{fig:retrieval}
	
\end{figure}

\subsection{Applications}
To further show the potential usage of our detection and description approach, we performed an evaluation in the two complementary tasks of object retrieval and 3D registration. 

\vspace*{0.2cm}\noindent\textbf{Deformable object retrieval.} 
We consider a database that contains images from various deformed objects. Each object appears multiple times with different deformations. The top K images from the database corresponding to an image query are retrieved. To evaluate the methods, we use the retrieval accuracy for different K values.
K-Nearest Neighbors is used in conjunction with Bag-of-Visual-Worlds approach over the descriptors as the retrieval engine. We compare our method against the state-of-the-art description methods that demonstrated the top performances in \cref{table:matching_score}, in addition to DELF~\cite{cit:DELF}, a state-of-the-art descriptor designed and trained specifically for image retrieval. We calculated the normalized area under the curve of each method for $K = \{1,...,20\}$. DALF achieved the most accurate retrieval capabilities at $99.49\%$, while DELF, DEAL, SuperPoint, and DISK achieved $98.57\%$, $98.34\%$, $97.92\%$, and $96.12\%$, respectively. \cref{fig:retrieval} shows some qualitative results\footnote{For the sake of clarity, we filtered out objects that all methods retrieved all correct images. Full results are available in the supplementary material.}.





	

\begin{table}[tb!]
	\centering
	\caption{{\bf 3D surface registration.} 
 The 2D and 3D accuracy is computed under varying thresholds in centimeters for the 3D residuals and in pixels for the 2D residuals. Best in bold.}
    \resizebox{0.99\columnwidth}{!}{%
	\begin{tabular}{lcccccc}
		\toprule      
		\multirow{2}{*}{\textbf{Method}} &\multicolumn{3}{c}{\textbf{2D Accuracy $\uparrow$}} & \multicolumn{3}{c}{\textbf{3D Accuracy $\uparrow$}}   \\ 
                                         & @2px & @3px & @5px     &      @0.5cm & @1.0cm & @1.5cm  \\
		\cmidrule(l){1-1}   \cmidrule(lr){2-4} \cmidrule(l){5-7}
        DISK     & 21.3 & 30.5 & 41.2   & 36.3 & 51.7 & 58.8 \\
        SuperPoint     & 23.2 & 34.4 & 47.4   & 42.6 & 60.5 & 68.4 \\
        SuperGlue     & 34.9 & 51.0 & \textbf{68.1}   & 42.8 & 64.4 & 73.6 \\
 	GeoPatch & 28.9 & 41.5 & 55.9   & 41.0 & 62.3 & 71.1 \\
        DEAL     & 29.4 & 42.0 & 56.2   & 42.9 & 64.3 & 72.8 \\       
        DALF (ours)     & \textbf{36.6} & \textbf{51.6} & 67.3   & \textbf{46.2} & \textbf{66.9} & \textbf{74.8} \\
		\bottomrule		
	\end{tabular}  
    }
	\label{table:arap}
	\vspace*{-0.3cm}
\end{table}

\paragraph{Non-rigid 3D surface registration.} 
 In addition to the retrieval application, we also validate the performance of the methods in the challenging real-world task of surface registration. To that end, we employ the as-rigid-as-possible (ARAP)~\cite{cit:ARAP} registration to perform the surface alignment with the correspondences obtained from each method. In these experiments, the correspondences are first filtered with an outlier removal approach~\cite{cit:adalam} since the ARAP cannot handle outliers in the registration process. One of the challenges of non-rigid registration is that one cannot fit a global geometric transformation with a minimal sample using RANSAC. 
 
 After the filtering stage, the ARAP is used to align the meshes of the respective image pairs. The 2D error is then computed using the ground-truth TPS transformation provided with the datasets given in pixels. We also estimate the residual 3D error, assuming that the two corresponding surfaces must be perfectly adjusted in the 3D space, as their meshes are known beforehand. 
 \cref{table:arap} shows the performance of the top methods under different thresholds for the 2D and the 3D errors considering all the datasets used in \cref{table:matching_score}, where our approach stands out, improving over $3$ p.p. in 3D registration accuracy compared to the best current method (SuperGlue) in the tightest threshold of $0.5$cm. Visual results of the 3D registration are available in the supplementary material.



\section{Conclusions} \label{sec:conclusion}
This paper presents DALF, a method that considers both the detection and description of keypoints under the challenging case of non-rigid geometric transformations. From extensive experiments and two applications using real deformable objects, we draw the following conclusions: (i) standard approaches for image matching deliver subpar results compared to deformation-aware features; (ii) optimizing the keypoint detection stage together with deformation-aware descriptors brings significant performance gains compared to existing deformation-aware methods that rely on affine keypoint detectors; and (iii) the feature fusion component is a simple but effective approach to increase the network expressiveness to deformations while keeping distinctiveness. 
\vspace{-0.3in}\paragraph{Acknowledgements.} The authors \new{would like to thank CAPES (\#88881.120236/2016-01), CNPq, FAPEMIG, and Google for funding different parts of this work. This work was also granted access to the HPC resources of CNRS IDRIS under the project 2021-AD011013154 and supported by the French Conseil Régional de Bourgogne-Franche-Comté.}






\balance
{\small
\bibliographystyle{ieee_fullname}

}

\newpage

\appendix

\title{[Supplementary Material] \\Enhancing Deformable Local Features by Jointly Learning to\\ Detect and Describe Keypoints }

\maketitle

In this supplementary material to our paper, we provide additional details about the retrieval and non-rigid 3D surface tracking applications, as well as further qualitative, quantitative results and visualization of the results presented in the paper. Please also have a look in our video showing detailed views of the non-rigid 3D surface registration for different approaches.

\section{Deformable object retrieval}
The nonrigid dataset contains various sequences of different objects being deformed over time. We selected one frame for each sequence to serve as a query image. The other frames of all sequences compose the search database, from where the application must retrieve the results. 

For each method, we detect and describe a maximum of $1{,}024$ keypoints inside a mask delimiting the object pixels. Next, we sample an equal amount of descriptors for each image to collect about $10{,}000$ descriptors. Then we use the sampled descriptors to compute $300$ centroids using the K-Means algorithm. The centroids are then used to calculate one global representation for each image using the Bag-of-Visual-Words approach to aggregate all the described keypoints. Given a query, we use the global descriptor to retrieve the closest $K$ images using $K$-Nearest Neighbors. We evaluate each method with the mean retrieval accuracy for each value of $K$ from $1$ to $20$.

\begin{figure}[h!]
    \centering
    \includegraphics[width=0.99\columnwidth]{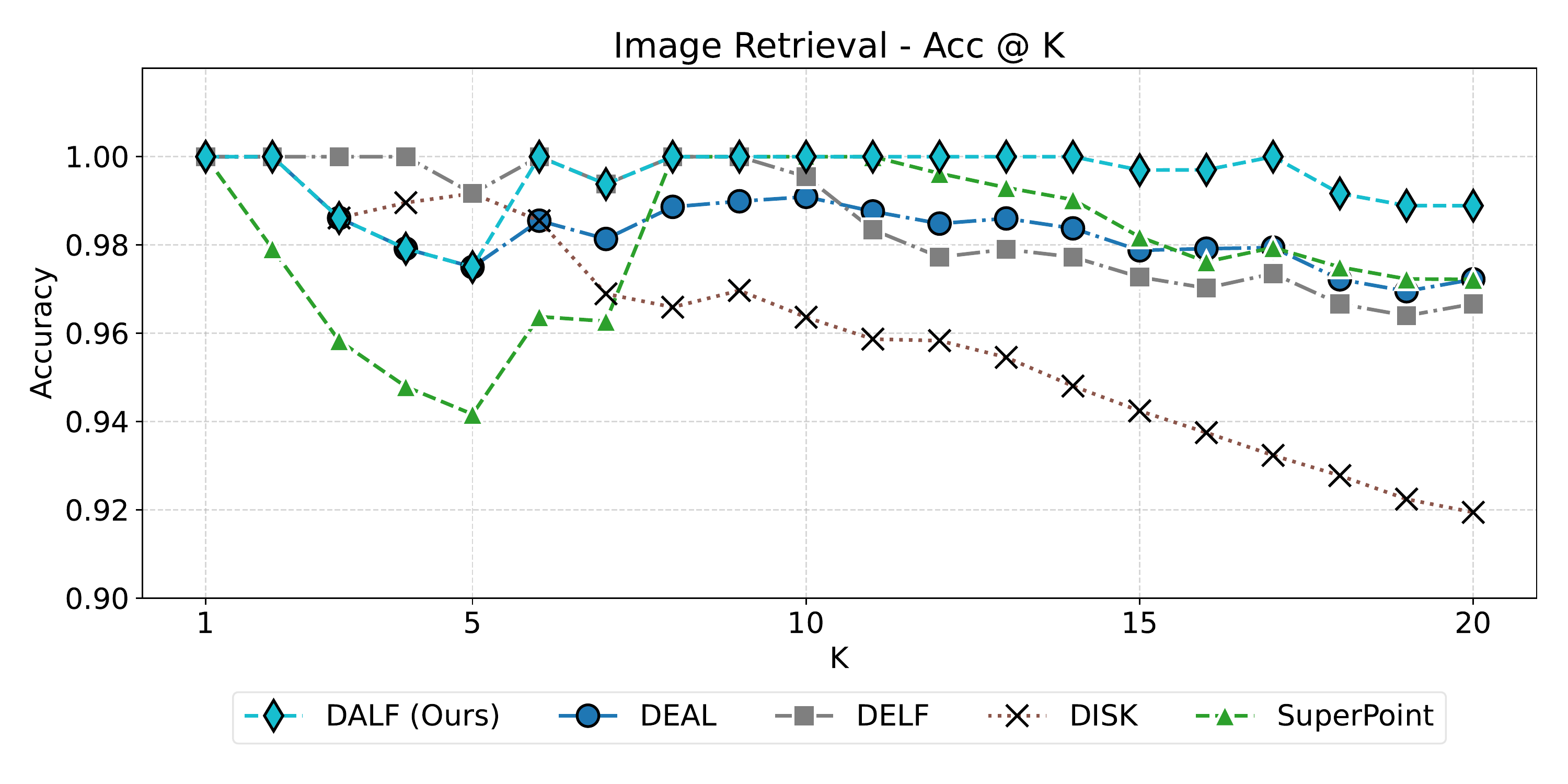}
    \caption{Accuracy@$K$ metric for the nonrigid object retrieval task. The normalized area-under-the-curve for each method is the following DISK: $96.12\%$, SuperPoint: $97.92\%$, DEAL: $98.34\%$. DELF: $98,57\%$, and DALF (Ours): $99,49\%$.}
    \label{fig:retrieval_quantitative}  
\end{figure}

\begin{figure*}[tb!]
	\centering
	\includegraphics[width=0.85\textwidth]{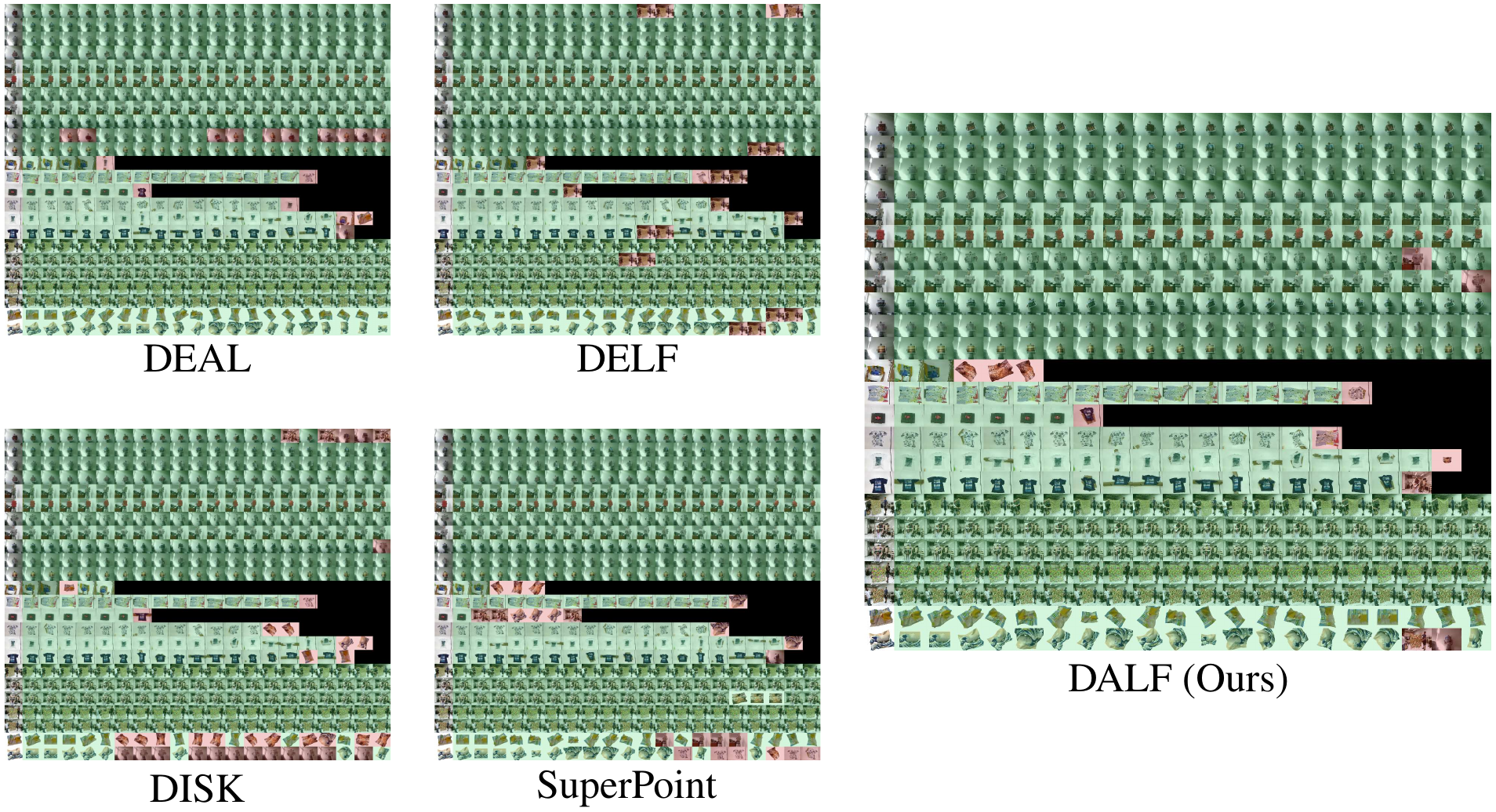}
	\caption{Our method has the best result in retrieving images of real and simulated deformed objects. The first column of each image shows the object queries, and the rows show the results from different queries. Green images correspond to the same object as the query, and red images do not correspond. Some objects are smaller and difficult to deform, so they may have less them $20$ occurrences in the dataset. In that case, we lower the value of $K$ to the exact number of occurrences of the object. For this reason, we can see some empty squares in the qualitative results. The black squares indicate no correspondent objects available.}
	\label{fig:retrieval_qualitative}
\end{figure*}

We compare our method against the best-performing description methods, in addition to DELF~\cite{cit:DELF}, a state-of-the-art descriptor designed and trained specifically for image retrieval. DALF achieved the best performance in the retrieval task, as shown in Figure \ref{fig:retrieval_quantitative}. \new{Note that at $ K=10 $, all methods achieve similar scores because they can correctly retrieve the easy images. However, note that the task becomes hard when $K > 10$, where all methods but DALF degrade as they cannot reliably retrieve the images of the heavily deformed objects, while DALF exhibits superior performance.} The full retrieval result for each query seen in Figure~\ref{fig:retrieval_qualitative}.  The code for the retrieval task will be publicly available; its objective is to be an easy-to-run benchmark for detectors and descriptors.

\new{ \section{Additional quantitative results}
In this section, we report additional metrics beyond the matching scores and mean matching accuracy, and present a more detailed analysis of the ablation study.

\paragraph{Keypoint Repeatability.}
Recent methods~\cite{cit:DISK, cit:aslfeat}, do not report keypoint repeatability because it often does not correlate well with downstream performance. Nevertheless, we computed repeatability across all the datasets, and our approach obtains the best average repeatability across all datasets. The scores are the following. DALF:~$0.58$, DISK:~$0.57$, non-rigid detector~\cite{cit:welerson2022}:~$0.47$, SIFT:~$0.41$, and R2D2:~$0.35$.

\paragraph{Extended ablation analysis.}
Although the two-stage training is not mandatory in our learning pipeline, it offers a better trade-off between invariance and distinctiveness, as shown in the top 3 performances on every dataset according to~\cref{table:extended_ablation} (which presents the scores per dataset from Tab. 2 of the main paper), thus we opt for C5 as the final design choice. Note that the fusion of the invariant and distinct features (C3--5), one of our novel contributions, achieves much better rankings on average across all datasets.
}

\begin{table}[t!]
\centering
\caption{\textbf{Extended ablation.} Matching score @ $3$ pixels for each configuration C following the order of Tab. 2 of the main paper, \eg, C1 corresponds to distinct-only, C2 to invariant-only, etc. Best result in \red{red}, second best in \green{green}, third best in \blue{blue}.}\vspace*{-7pt}
\resizebox{0.80\columnwidth}{!}{%
\footnotesize
\begin{tabular}{lccccc}
\toprule
\textbf{Dataset} & \textbf{C1} & \textbf{C2} & \textbf{C3} & \textbf{C4} & \textbf{C5}  \\
\midrule

\textit{Kinect1} & $\red{0.58}$  & $0.52$  & $0.53$  & $\green{0.55}$  & $\blue{0.54}$ \\ 
\textit{Kinect2} & $0.54$  & $0.56$  & $\blue{0.60}$  & $\red{0.63}$  & $\green{0.62}$ \\ 
\textit{DeSurT}  & $\blue{0.48}$  & $0.43$  & $0.46$  & $\red{0.50}$  & $\green{0.49}$ \\ 
\textit{Simulation} & $0.27$  & $\red{0.52}$  & $\green{0.50}$  & $0.34$  & $\blue{0.42}$ \\ 
\bottomrule
\vspace*{-22pt}
\end{tabular}
}
\label{table:extended_ablation}
\end{table}

\section{Non-rigid 3D surface registration}
In this section, we describe in detail the implementation of the surface registration application using the as-rigid-as-possible (ARAP) ~\cite{cit:ARAP} optimization, and also show qualitative results derived from the experiments of Tab. 3 (3D surface registration) of the paper.

\begin{figure*}[tb!]
	\centering
	\includegraphics[width=1.7\columnwidth]{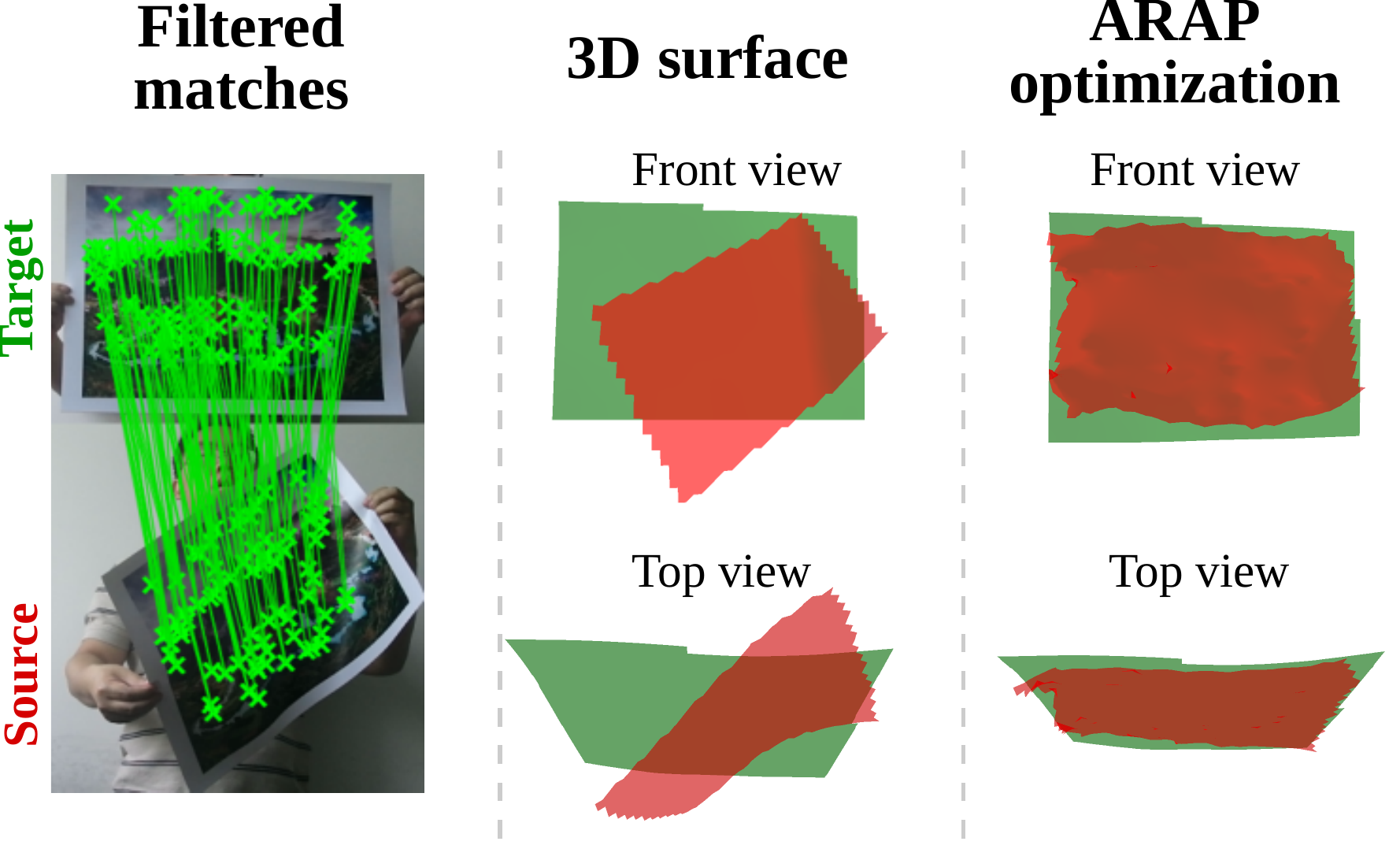}
	\caption{\textbf{Non-rigid 3D surface registration overview}. We use the filtered correspondences (left) to align two meshes of the same surface obtained from their respective RGB-D frames (middle) to the same reference pose and deformation (right), using as-rigid-as-possible (ARAP) refinement. }

	\label{fig:teaser_sup}  \vspace{-0.11in}
\end{figure*}

Non-rigid 3D surface registration aims to accurately align two RGB-D frames of the same surface, viewed from different viewpoints at the same time that the object is affected by non-rigid deformations. \Cref{fig:teaser_sup} shows an overview of the registration pipeline. Surface alignment is a crucial step used by non-rigid reconstruction frameworks~\cite{cit:volumedeform, bozic2020deepdeform} that allow complete 3D reconstruction of deforming objects. Improvements in registration accuracy can significantly increase the quality of the reconstruction, enabling the use of such systems in critical, challenging applications, such as the live reconstruction of human organs~\cite{cit:laparoscopy}.

\subsection{Implementation details}

 \begin{figure*}[tb!]
	\centering
	\includegraphics[width=0.99\textwidth]{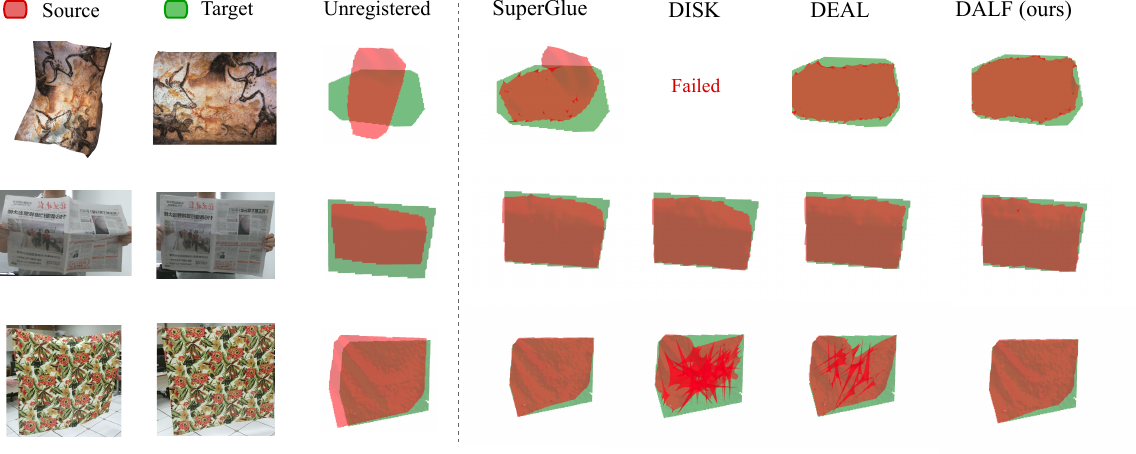}
	\caption{\textbf{Non-rigid registration under challenging scenarios}. Our method can achieve accurate non-rigid registration under large rotations, illumination changes caused by deformations, and highly repetitive patterns. In contrast, all other techniques produce low-quality results in at least one of the challenging scenarios. The sharp line artifacts in two registrations from DEAL and DISK indicates that the method produced inconsistent matches even after the filtering step, and the ARAP optimization failed due to local minima. Please check the supplementary video to visualize the registration results in 3D with the depicted image pairs and additional samples.}

	\label{fig:qualitative-reg}  
\end{figure*}

Our application considers the most difficult scenario: wide-baseline registration, where the object can be in an arbitrary viewpoint and deformed shape. Thus, it is challenging to filter outlier matches, in contrast with rigid registration, where it is possible to fit a homography or fundamental matrix using a minimal correspondence sample and perform RANSAC to remove the outlier correspondences with high confidence. 
 
Our solution to this problem was to tune the AdaLAM~\cite{cit:adalam} filtering method to perform outlier detection in the presence of image deformations. AdaLAM checks the affine consistency of local point clusters and filters the correspondences that are inconsistent with their neighboring matches. As we have observed empirically, the assumption of localized affine consistency is a reasonable approximation for non-rigid correspondences. We adjusted the sensitivity of the local affine RANSAC of AdaLAM to tolerate more deviation from the base affine transformation, which usually happens in the presence of significant deformations. 

AdaLAM tends to provide erroneous consistent affine matches when the scene has repetitive patterns, which is inevitable in practice. Those inconsistent matches introduce large errors in the ARAP optimization, and the method fails to return a meaningful result. Thus, to improve the robustness of the registration, for all methods, we use the best $200$ matches according to the Lowe's ratio test~\cite{lowe2004ijcv}, which drastically reduces artifacts caused by repetitive patterns, and also accelerates the convergence of the ARAP optimization. The non-rigid registration application source-code will be released alongside the reference implementation of our proposed method.

\subsection{Qualitative results}
\cref{fig:qualitative-reg} shows reconstruction results of challenging samples from the non-rigid datasets, where our method obtains robust registration. Aside from this PDF document, we made available a video displaying the rendered registered surfaces in 3D from our approach and the competing methods (please visit~\url{verlab.dcc.ufmg.br/descriptors/dalf_cvpr23} to see the video), where it is possible to visualize the registration quality better.
It is worth mentioning that SuperGlue~\cite{superglue}, the best competing method, requires inputs in the form of image pairs, and employs global self and cross attention across local features when matching them , \ie, the matching problem is conditioned to the input image pair, which significantly improves its robustness, especially in ambiguous regions. In contrast, our method independently detects the features, and a simple nearest neighbor search is used to perform matching. Our strategy can empower SuperGlue with deformation awareness by simply using our descriptors during training. In turn, SuperGlue's global self and cross-attention mechanisms can help our approach become much more robust to matching in challenging scenarios.

\balance

\end{document}